\documentclass[journal]{IEEEtran}
\usepackage{amsmath,amsfonts,bm}

\def\figref#1{figure~\ref{#1}}
\def\secref#1{section~\ref{#1}}
\def\eqref#1{equation~\ref{#1}}
\def\1{\bm{1}}

\def\mF{{\bm{F}}}

\def\mW{{\bm{W}}}
\def\mX{{\bm{X}}}
\def\mY{{\bm{Y}}}
\def\mZ{{\bm{Z}}}

\DeclareMathAlphabet{\mathsfit}{\encodingdefault}{\sfdefault}{m}{sl}
\SetMathAlphabet{\mathsfit}{bold}{\encodingdefault}{\sfdefault}{bx}{n}

\newcommand{\R}{\mathbb{R}}

\usepackage{graphicx, overpic}
\usepackage{amsmath}
\usepackage{amssymb}
\usepackage[symbol]{footmisc}
\usepackage[ruled,norelsize]{algorithm2e}
\usepackage{setspace}
\usepackage[thinlines]{easytable}
\usepackage{array,multirow,booktabs}
\usepackage{diagbox}
\usepackage{makecell}
\usepackage{xcolor, colortbl}
\usepackage{arydshln}
\usepackage{cite}
\usepackage[bookmarks=true]{hyperref}
\hypersetup{
    breaklinks=true,
    citecolor=blue,
    linkcolor=red,
    urlcolor=blue,
    colorlinks=true,
}

\makeatletter
\newcommand{\removelatexerror}{\let\@latex@error\@gobble}
\makeatother
\renewcommand{\eqref}[1]{Eq.~(\ref{#1})}
\renewcommand{\figref}[1]{Fig.~\ref{#1}}%
\newcommand{\tabref}[1]{Tab.~\ref{#1}}%
\renewcommand{\secref}[1]{Sec.~\ref{#1}}

\def\eg{\textit{e.g.,}}
\def\ie{\textit{i.e.,}}
\def\etal{\textit{et al.}}

\def\Uspace{\vspace{0pt}}
\def\Lspace{\vspace{0pt}}
\def\PARAspace{\vspace{5pt}}

\RequirePackage{silence}
\begin{document}

\title{Dependency Aware Filter Pruning}

\author{Kai~Zhao, Xin-Yu~Zhang, Qi~Han, 
  and Ming-Ming~Cheng% <-this % stops a space
  \thanks{The first three students make equal contributions to this paper.}
  \thanks{K. Zhao, XY. Zhang, Q. Han, and MM. Cheng are with
  TKLNDST, CS, Nankai University, Tianjin, China.}
}

\markboth{Journal of \LaTeX\ Class Files,~Vol.~14, No.~8, August~2015}%
{Shell \MakeLowercase{\textit{et al.}}: Bare Demo of IEEEtran.cls for IEEE Journals}

\maketitle

\begin{abstract}
Convolutional neural networks (CNNs) are typically over-parameterized,
bringing considerable computational overhead and memory footprint in inference.
Pruning a proportion of unimportant filters is an efficient way to
mitigate the inference cost.
For this purpose, identifying unimportant convolutional filters is the key to
effective filter pruning.
Previous work prunes filters according to either their weight norms
or the corresponding batch-norm scaling factors,
while neglecting the sequential dependency between adjacent layers.
In this paper, we further develop the norm-based importance estimation by taking the
dependency between the adjacent layers into consideration.
Besides, we propose a novel mechanism to dynamically control the sparsity-inducing regularization
so as to achieve the desired sparsity.
In this way, we can identify unimportant filters and search for the optimal network
architecture within certain resource budgets in a more principled manner.
Comprehensive experimental results demonstrate the proposed method performs
favorably against the existing strong baseline on the CIFAR, SVHN, and ImageNet datasets.
The training sources will be publicly available after the review process.
\end{abstract}

\begin{IEEEkeywords}
Deep Learning, Nertwork Compression, Filter Pruning.
\end{IEEEkeywords}

\IEEEpeerreviewmaketitle

\Uspace
\section{Introduction}
\Lspace

Convolutional neural networks (CNNs) have achieved remarkable performance
on a wide range of vision and learning tasks
~\cite{krizhevsky2012imagenet,deshpande2015fully,girshick2015fast,long2015fully,sun2014deep,yang2014multitask,chang2016bi,
luo2017adaptive,du2016stacked,wei2016cross,zhang2016visual}.
% among low level vision,
% such as edge detection~\cite{xie2015holistically},
% and high level vision, including semantic segmentation~\cite{long2015fully}
% and detection~\cite{girshick2015fast}.
%
Despite the impressive performance,
CNNs are notably over-parameterized and thus lead to
high computational overhead and memory footprint in inference.
Therefore, network compression techniques are developed to assist
% Therefore, network compression techniques are beneficial to
the deployment of CNNs in real-world applications.

\begin{figure}[!tb]
  \centering
  \begin{overpic}[width=1\linewidth]{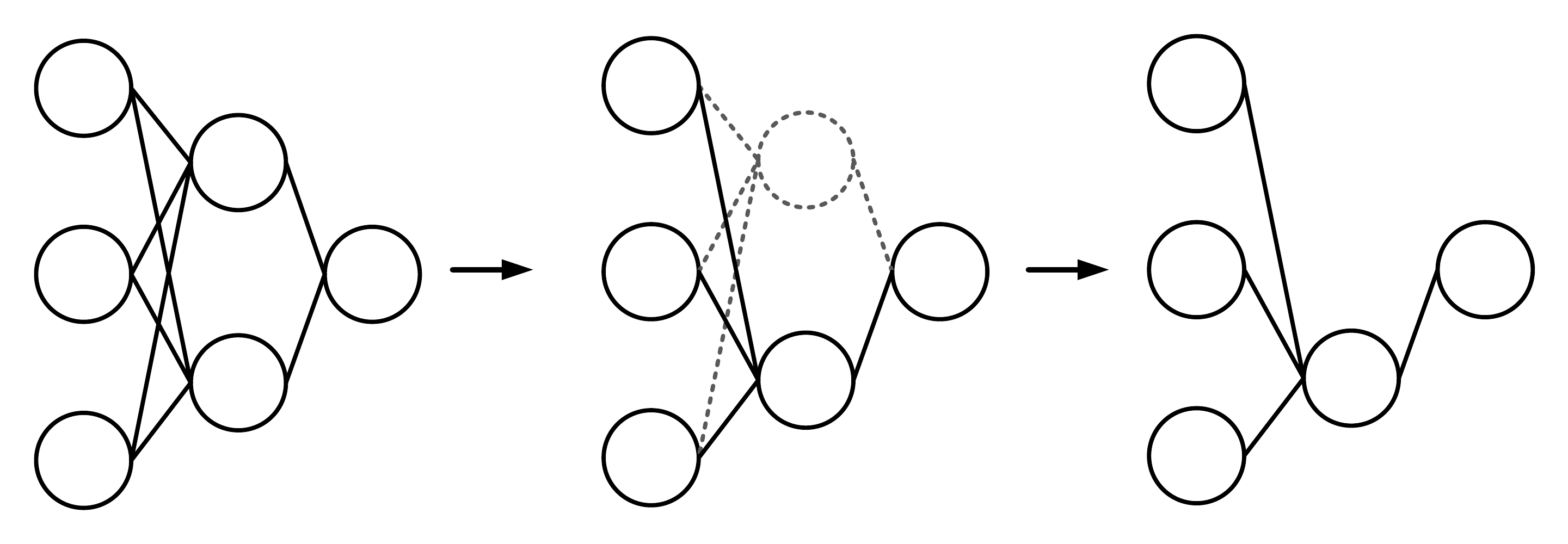}
  \put(10,-3){(a)}
  \put(45,-3){(b)}
  \put(80,-3){(c)}
  \end{overpic}
  \vspace{1pt}
  \caption{A typical pipeline of filter pruning:
  (a) train a over-parameterized model with sparsity-inducing regularization;
  (b) prune unimportant filters based on certain criteria;
  (c) finetune the compressed model till convergence.
  }\label{fig:pipeline}
\end{figure}

Filter pruning is an efficient way to reduce
the computational cost of CNNs with negligible performance degradation.
As shown in \figref{fig:pipeline}, a typical pipeline of filter pruning
\cite{liu2017learning} works as follows:
1) train an over-parameterized model with the sparsity-inducing regularization;
2) estimate the importance of each filter and prune the unimportant filters;
3) finetune the compressed model to recover the accuracy.
Among these, identifying unimportant filters is the key to efficient
filter pruning.
Prior work \cite{gordon2018morphnet,li2017pruning,liu2017learning,wen2016learning}
prunes filters according to the magnitude of the corresponding model parameters.
For example, Li~\etal~\cite{li2017pruning} prune convolutional filters of smaller
$L_1$ norms as they are considered to have less impact on the functionality of the network.
Network Slimming~\cite{liu2017learning} then proposes to prune channels (\ie~filters)
based on the corresponding scaling factors.
To be specific, the scaling factors of the batch normalization (BN)
\cite{ioffe2015batchbatch} layer serve as an indicator of the channel importance,
on which an $L_1$ regularization is imposed to promote sparsity.
%
% Channels with smaller scaling factors are pruned from the network.
% \cite{liu2017learning} adopts the scaling factors in batch normalization
% (BN) \cite{ioffe2015batchbatch} layers as the indicator of channel importance,
% and channels with smaller scaling factors are pruned from the network.
%
% Furthermore, an $L_1$ penalty is imposed on the scaling factors to promote sparsity.
%
As a result, Liu \etal~\cite{liu2017learning} derive an automatically searched
network architecture of the compressed model.
%
% Often a sparse regularization is imposed either to the model parameters~\cite{li2017pruning}
% to derive sparse models.
%
% Some methods~\cite{he2017channel,wen2016learning} exploit a LASSO regression to figure out redundant
% filters.
%
% Network slimming~\cite{liu2017learning} proposes to prune less important filters according to the
% magnitude of scale factors of batchnorm~\cite{ioffe2015batchbatch} layers.
%
% Then convolution filters with smaller scale factors are pruned out from the model.
%
% This method derives a automatically selected network architecture.

However, existing methods select unimportant filters based only on the parameter magnitude of
a single layer~\cite{gordon2018morphnet,li2017pruning,liu2017learning,he2018soft,he2019asymptotic,ye2018rethinking},
while neglecting the dependency between consecutive layers.
For example, a specific channel with a small BN scaling factor may be
followed by a convolution with a large weight magnitude at that channel,
making the channel still important to the output.
% For example, feature maps with small BN factors
% may correspond to a large convolutional filter in the next layer, making these feature maps
% still important to the output.
%
Besides, in the ``smaller BN factor, less importance'' strategy, BN factors from
different layers are gathered together to rank and determine the filters to be pruned.
We argue and empirically verify that this strategy is sub-optimal and may lead to
unstable network architectures as it neglects 
the intrinsic statistical variation among the BN factors of different layers.
Empirically, we observe that the pruned architectures of Network Slimming
\cite{liu2017learning} are sometimes unbanlanced and lead to severely degraded performance,
especially when the pruning ratio is relatively high.

In this paper, we propose a \textit{dependency-aware filter pruning} strategy,
which takes the relationship between adjacent layers into consideration.
Hence, we measure the filter importance in a more principled manner.
% so as to approximate the filter importance more accurately.
%
Along this line, we introduce a novel criteria to determine the filters to be pruned
by the local importance of the consecutive two layers.
That is, if one layer is sparse, then more filters will be pruned and vice versa,
regardless of the statistics of other layers.
% In our method, the filter selection is made independent of layers based on one layer's
% local statistics: if a layer is sparse, there will be more filters pruned and vice versa.
%
Finally, we propose an automatic-regularization-control mechanism in which the coefficient
of the sparsity-inducing regularization is dynamically adjusted to meet the desired sparsity.
% Finally, we propose to automatically control the coefficient of the sparsity regularization
% to achieve the desired model sparsity.
%
Our contributions are summarized below:
\begin{itemize}
    \item We propose a principled criteria of measuring the filter importance
    by taking the dependency between adjacent layers into consideration.

    \PARAspace
    \item Given the dependency-aware filter importance, we prune filters
    based on the local statistics of each layer,
    instead of ranking the filter importance across the entire network.

    \PARAspace
    \item We propose to dynamically control the coefficient of the sparsity-inducing
    regularization to achieve the desired model sparsity.
\end{itemize}
Comprehensive experimental results demonstrate that the improved filter pruning strategy
performs favorably against the existing strong baseline \cite{liu2017learning} on the
CIFAR, SVHN, and ImageNet datasets.
We also validate our design choices with several ablation studies and verify that
the proposed algorithm reaches more stable and well-performing architectures.

\Uspace
\section{Related work}
\Lspace

\subsection{Network pruning}
\Lspace
Network pruning is a prevalent technique to reduce redundancy in deep neural networks
by removing unimportant neurons.
Specifically, \textit{weight pruning} approaches
\cite{carreira2018learning,dong2017learning,guo2016dynamic,han2015learning,hassibi1993second,
lecun1990optimal,srinivas2017training}
remove network parameters without structural constraints,
thus leading to unstructured architectures that are not well supported by the BLAS libraries.
On the other hand, \textit{filter pruning} methods
\cite{li2017pruning,liu2017learning,he2019filter,he2017channel,molchanov2019importance,zeng2018accelerating}
remove the entire filters (\ie~channels) from each layer,
thus resulting in compact networks that can be conveniently incorporated into modern BLAS libraries.
According to how to identify the unimportant filters,
existing filter pruning methods can be further divided into two categories:
\textit{data-dependent filter pruning} and \textit{data-independent filter pruning}.

Data-dependent filter pruning utilizes the training data
to determine the filters to be pruned.
Polyak ~\etal~\cite{Polyak2015Channel} remove filters that
produce activations of smaller norms.
He ~\etal~\cite{he2017channel} perform a channel selection
by minimizing the reconstruction error.
Zheng~\etal~\cite{Zheng2015Compact} and Anwar~\etal~\cite{anwar2017structured}
both evaluate the filter importance via the loss of the validation accuracy without each filter.
Molchanov ~\etal~\cite{molchanov2019importance} approximate the exact contribution
of each filter with the Taylor expansion.
A recent work \cite{zhou2018accelerate} proposes a layer-wise recursive Bayesian pruning method
with a dropout-based metric of redundancy.

Data-independent filter pruning identifies less important filters based
merely on the model itself (\ie~model structure and model parameters).
%
% The exact formulation of pruning is a $L_0$
% minimization problem under performance constraints.
% %
% However, $L_0$ minimization is NP-hard problem
% due to the non-convexity of the $L_0$ norm.
% %
% Therefore, researchers relax the $L_0$ minimization
% with $L_1$ or $L_2$ regularization, and prune parameters
% whose magnitudes are below a pre-defined threshold.
% %
% For example, Han \etal~\cite{han2015learning} 
% %
Li \etal~\cite{li2017pruning} discard filters according to
the $L_1$ norm of the corresponding parameters as filters
with smaller weights are considered to contribute less to the output.
Network Slimming~\cite{liu2017learning} imposes a sparsity-inducing regularization on
the scaling factors of the BN layer and then prunes filters with smaller scaling factors.
Zhou~\cite{zhou2019knee} using the evolutionary algorithm to search redundant filters during training.
He~\etal~\cite{he2019asymptotic} propose to dynamically prune filters
during training.
In another work He~\etal~\cite{he2019filter} propose to prune filters
that are close to the geometric median.
They argue that filters near the geometric median are more likely to be
represented by others \cite{fletcher2008robust}, thus leading to redundancy.
%

%of all the filters in the same layer.

Our method belongs to the \textit{data-independent filter pruning},
which is generally more efficient as involving the training data brings extra computation.
For example, Zheng~\etal~\cite{Zheng2015Compact} and Anwar~\etal~\cite{anwar2017structured}
measure the importance of each filter by removing the filter
and re-evaluating the compressed model on the validation set.
This procedure is extremely time-consuming.
Essentially, we take the dependency between the consecutive layers into consideration,
while previous data-independent methods~\cite{li2017pruning,liu2017learning,he2019filter}
merely focus on the parameters
(either the convolutional weights~\cite{li2017pruning,he2019filter}
or the BN scaling factors~\cite{liu2017learning})
of a single layer.
Besides, we propose a novel mechanism to dynamically control the coefficient
of the sparsity-inducing regularization,
instead of pre-defining it based on human heuristics \cite{liu2017learning}.
Incorporating these components, our principled approches and better estimate
the filter importance (\secref{sec:importance-estimation}) and
achieve more banlanced pruned architectures (\secref{sec:prune-stable}).
% Extensive experimental results demonstrate the superiority of our principled pruning strategy.

\Uspace
\subsection{Neural Architecture Search}
\Lspace

While most state-of-the-art CNNs \cite{he2016deep,huang2017densely,simonyan2014very}
manually designed by human experts, there is also a line of research that explores
automatic network architecture learning
\cite{cai2018proxylessnas,dai2019chamnet,liu2018progressive,pham2018efficient,
tan2019mnasnet,wu2019fbnet,zoph2016neural},
called neural architecture search (NAS).
Specifically, automatically tuning channel width is also studied in NAS.
For example, ChamNet \cite{dai2019chamnet} builds an accuracy predictor on
the Gaussian Process with the Bayesian optimization
to predict the network accuracy with various channel widths in each layer.
FBNet \cite{wu2019fbnet} adopts a gradient-based method to optimize the CNN
architecture and search for the optimal channel width.
The proposed pruning method can be regarded as a particular case of
channel width selection as well, 
except that we impose the resource constraints on the selected architecture.
However, our method learns the architecture through a single training process,
while typical NAS methods may train hundreds of models with different architectures
to determine the best-performing one \cite{dai2019chamnet,zoph2016neural}.
We highlight that our efficiency is in line with the goal of neural architecture search.

\begin{figure*}[!tb]
  \centering
  \begin{overpic}[width=1\linewidth]{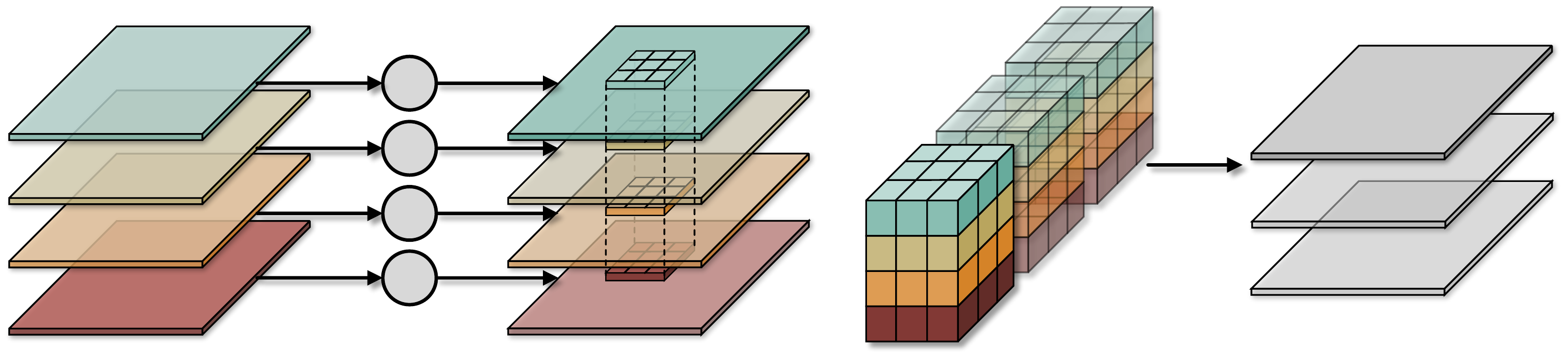}
    \put(25,17){$\gamma_4^l$}
    \put(25,12.9){$\gamma_3^l$}
    \put(25,8.7){$\gamma_2^l$}
    \put(25,4.6){$\gamma_1^l$}
    \put(8,22.3){\footnotesize{feature maps}}
    \put(2,18.3){\footnotesize{$\mX^l$}}
    %\put(23,20.2){\footnotesize{BN factors}}
    \put(35,19.3){\footnotesize{$\mY^l $}}    
    \put(37,22.3){\footnotesize{scaled feature maps}}
    \put(67,23.3){\footnotesize{Conv kernels}}
    % \put(84.5,20.9){\footnotesize{output feature maps}}
    \put(74.1,20.6){\tiny{$||\widetilde{\mW}_4^{l+1}||$}}
    \put(74.1,16.6){\tiny{$||\widetilde{\mW}_2^{l+1}||$}}
    \put(53.0, 12.5){\large{$\circledast$}}
  \end{overpic}
  \caption{
    Illustration of the ``batch normalization, convolution'' sequential.
    % A batch normalization $\rightarrow$ convolution sequential.
    %
    The input feature $\mX^l$ (after whitening) is stretched by the scaling factors
    $\gamma^l$ and convolved with three convolutional kernels,
    resulting in a three-channel output feature.
    For the dependency-aware criterion, the importance of the $c^{th}$ feature map is
    estimated by the product of the absolute value of the scaling factor $|\gamma_c^l|$ and
    the magnitude of corresponding convolutional kernel $||\widetilde{\mW}_c^{l+1}||$.
    (See \eqref{eq:channel-importance}.)
   }
   \label{fig:dep-prune}
\end{figure*}
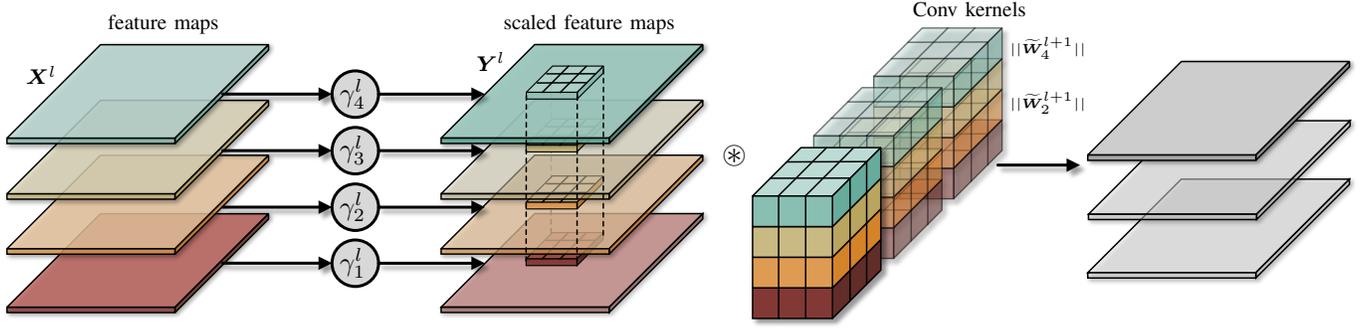

\Uspace
\subsection{Other Alternatives for Network Compression}
\Lspace

\paragraph{Low-Rank Decomposition}
There is a line of research
\cite{denton2014exploiting,sindhwani2015structured,zhang2015accelerating,huang2018ltnn}
that aims to approximate the weight matrices of the neural networks
with several low-rank matrices using techniques like the
Single Value Decomposition (SVD) \cite{denton2014exploiting}.
However, these methods cannot be applied to the convolutional weights,
and thus the acceleration in inference is limited.

\PARAspace
\paragraph{Weight Quantization}
Weight quantization \cite{chen2015compressing,courbariaux2016binarized,
rastegari2016xnor,wu2016quantized,cheng2017quantized}
reduces the model size by using a low bit-width number
of the weights and hidden activations.
For example, Courbariaux \etal~\cite{courbariaux2016binarized}
and Rastegari \etal~\cite{rastegari2016xnor}
quantize the real-valued weights into binary or ternary ones,
\ie~the weight values are restricted to $\{-1, 1\}$ or $\{-1, 0, 1\}$.
Cheng~\etal~\cite{cheng2017quantized} quantize CNNs with a predefined codebok.
Despite the significant model-size reduction and inference acceleration,
these methods often come with a mild accuracy drop due to the low precision.
%--------------------------------------------------------------------------------------------------------------------%
%--------------------------------------------------------------------------------------------------------------------%
\Uspace
\section{Dependency-Aware Filter Pruning}
\Lspace

%--------------------------------------------------------------------------------------------------------------------%
\subsection{Dependency Analysis} \label{sec:dependency}
\Lspace

Generally, we assume a typical CNN involves multiple convolution operators (Conv layers), batch normalizations
(BN layers) \cite{ioffe2015batchbatch}, and non-linearities,
which are applied to the input signals sequentially as in \figref{fig:dep-prune}.
Practically, each channel is transformed independently in the BN layers and non-linearities,
while inter-channel information is fused in the Conv layers.
To prune filters (\ie~channels) with minimal impact on the network output,
we analyze the role each channel plays in the Conv layers as follows.

Let $\mX^l \in \R^{C^l \times H^l \times W^l}$ be the hidden activations after normalization before scaling in the
$l^{th}$ BN layer.
The scaled activations $\mY^l$ can be formulated as\footnote{For simplicity, we omit the shifting parameters in a
typical BN layer, and the bias term in \eqref{equ:conv}.}
\begin{equation}
    \mY_c^l = \gamma_c^l \mX_c^l,
\end{equation}
where $\gamma^l \in \R^{C^l}$ denotes the scaling factor of the $l^{th}$ BN layer
and $\mX_c^l$ (\textit{resp.} $\mY_c^l$) is the $c^{th}$ channel of $\mX^l$ (\textit{resp.} $\mY^l$).
Then, a Lipschitz-continuous non-linearity $\sigma$ is applied to $\mY^l$, namely,
\begin{equation}
    \mZ^l = \sigma(\mY^l).
\end{equation}
Afterward, all channels of $\mZ^l$ are fused into $\mF^{l+1} \in \R^{C^{l+1} \times H^{l+1} \times W^{l+1}}$
via a convolution operation,
and different channels contribute to the fused activation $\mF^{l+1}$ differently.
Formally, let $\mW^{l+1} \in \R^{C^{l+1} \times C^{l} \times k \times k}$ be the $(l+1)^{th}$ convolution filter,
where $k$ denotes the kernel size.
We have 
\begin{equation}
    \mF^{l+1} = \mW^{l+1} \circledast \mZ^{l},\label{equ:conv}
\end{equation}
where $\circledast$ denotes the convolution operator.
As convolution is an affine transformation, we re-formulate the linearity of \eqref{equ:conv} explicitly:
\begin{equation}
    \widetilde{\mF}^{l+1} = \widetilde{\mW}^{l+1} \widetilde{\mZ}^{l},
\end{equation}
where $\widetilde{\mF}^{l+1} \in \R^{C^{l+1} \times H^{l+1}W^{l+1}}$, $\widetilde{\mW}^{l+1} \in \R^{C^{l+1} \times k^2C^{l}}$,
and $\widetilde{\mZ}^{l} \in \R^{k^2C^{l} \times H^{l+1}W^{l+1}}$ are the unfolded versions of $\mF^{l+1}$, $\mW^{l+1}$,
and $\mZ^{l}$, respectively.
Factorize $\widetilde{\mF}^{l+1}$ along the channel axis, and we have
\begin{equation}
    \widetilde{\mF}^{l+1} = \sum_{c=1}^{C^{l}} \widetilde{\mW}_c^{l+1} \widetilde{\mZ}_c^{l},
\end{equation}
where $\widetilde{\mW}_c^{l+1} \in \R^{C^{l+1} \times k^2}$ and $\widetilde{\mZ}_c^{l} \in \R^{k^2 \times H^{l+1}W^{l+1}}$.
Then, we analyze the contribution of each channel as follows:\footnote{Here, we assume the non-linearity provides
zero activations given zero inputs, and most widely-used non-linearities, such as ReLU \cite{nair2010rectified} and
its variants~\cite{maas2013rectifier,clevert2016fast,he2015delving}, satisfy this property.}

\begin{align}
    ||\widetilde{\mF}^{l+1}|| & \leq \sum_{c=1}^{C^{l}} ||\widetilde{\mW}_c^{l+1} \widetilde{\mZ}_c^{l}||
    \leq \sum_{c=1}^{C^{l}} ||\widetilde{\mW}_c^{l+1}|| \cdot ||\widetilde{\mZ}_c^{l}|| \notag \\
    & \leq \sum_{c=1}^{C^{l}} ||\widetilde{\mW}_c^{l+1}|| \cdot \mathcal{L} ||\widetilde{\mY}_c^{l}|| \notag \\
    & = \mathcal{L} \sum_{c=1}^{C^{l}} |\gamma_c^{l}| \cdot ||\widetilde{\mW}_c^{l+1}|| \cdot ||\widetilde{\mX}_c^{l}||,
\end{align}
where $\mathcal{L}$ denotes the Lipschitz constant of function $\sigma$, and $\widetilde{\mX}^{l}$ and $\widetilde{\mY}^{l}$ are the unfolded versions of $\mX^{l}$ and $\mY^{l}$,
respectively.
Since the normalization operation in BN layer uniformize the activations $\mX_c^{l}$ (\ie~$\widetilde{\mX}_c^{l}$)
across channels,
we quantify the contribution of the $c^{th}$ channel by
\begin{equation}
  S_c^{l} = |\gamma_c^{l}| \cdot ||\widetilde{\mW}_c^{l+1}||,
  \label{eq:channel-importance}
\end{equation}
which serves as our metric for network pruning.

\Uspace
\subsection{Filter Selection}\label{sec:filter-selection}
\Lspace

Let $r \in (0, 1)$ be the pruning ratio, and $C^l~(l \in \{1,2,\cdots,L\})$ be the number of filters in the $l^{th}$
convolutional layer.
Generally, previous works can be divided into two groups according to the target network.

\PARAspace
\paragraph{Pruning with Pre-defined Target Network}
Many previous work~\cite{he2018soft,he2019asymptotic,he2019filter} prune a fixed ratio of filters in each layer.
In other words, there will be $r\cdot C^l$ filters pruned from the $l^{th}$ layer.
The architecture of the target network is known even without pruning.
However, recent work~\cite{gordon2018morphnet,li2019data} reveals
that this stretagy cannot find the optimal distribution of the neuron numbers
of each convolutional layer across the network,
as some layers will be over-parameterized while some under-parameterized.

\PARAspace
\paragraph{Pruning as Architecture Search}
Network Slimming~\cite{liu2017learning} treats pruning as a special form of architecture search,
\ie~search for the optimal channel width of each layer.
It compares the importance of each convolutional filter across the entire network
and prunes filters of less importance.
This approach provides more flexibility of the compressed architecture
as a higher pruning ratio can be achieved if a specific layer is sparse
and vice versa.
% can be regarded as an automatic architecture search because sparser layers
% will have more filters pruned.

However, according to our practice,
we find that sometimes too many filters of a layer
(or occasionally all filters of a layer) are pruned in this strategy,
leading to severely degraded performance.
This is because it does not take the intrinsic statistical variation
among different layers into consideration.
Suppose there are two layers and the corresponding scaling factors are
$\{0.10, 0.01, 0.03, 0.15\}$ and $\{1, 100, 2, 200\}$, respectively.
Our target is to prune half of the filters, \ie~$r=0.5$.
Apparently, the second and third channels should be pruned from the first layer,
and the first and third channels should be pruned from the second layer.
However, if we rank the scaling factors globally, all filters of the first
layer will be pruned, which is obviously unreasonable.

To alleviate this issue, we instead select the unimportant filters
based on the intra-layer statistics.
Let $S^l_c$ be the importance of the $c^{th}$ channel in the $l^{th}$ layer.
Then, filters with importance factor $S^l_c \le \max(S^l)\cdot p$ will be pruned,
where the threshold $p \in (0, 1)$ is a hyper-parameter.
Formally, the set of filters to be pruned in the $l^{th}$ layer is:
\begin{equation}
  \mathcal{F}_{\text{pruned}}^l= \{ c : S^l_c \le \max(S^l)\cdot p\}.
  \label{eq:filters-pruned}
\end{equation}
In our solution, the choice of the filters to be pruned in one layer
is made independent of the statistics of other layers, 
so that the intrinsic statistical differences among layers will not
result in dramatically unbalanced neural architecture.

%--------------------------------------------------------------------------------------------------------------------%
\Uspace
\subsection{Automatic Control of Sparsity Regularization} \label{sec:auto-control}
\Lspace

Network Slimming \cite{liu2017learning} imposes an $L_1$ regularization
on the model parameters to promote model sparsity.
However, choosing a proper regularization coefficient $\lambda$
is non-trivial and mostly requires manual tuning based on human heuristics.
%
% A large pruning ratio $r$ requires a high degree of
% model sparsity and thus, a large $\lambda$.
%
% While the existence of $\rm L1$ regularization will do harm to the optimization,
% leading to sub-optimal performance.
%
For example, Network Slimming performs a grid search in a set of candidate
coefficients for each dataset and network architecture.
However, different pruning ratios require different levels of model sparsity,
and thus different coefficients $\lambda$.
It is extremely inefficient to tune $\lambda$ for each experimental setting.
%
% Given the pruning scheme in \secref{sec:filter-selection},
% one cannot directly determine the total number of filters to be pruned.

To escape from manually choosing $\lambda$ and meet the required model sparsity at the same time,
% number of filters to be pruned at the same time,
we propose to automatically control the regularization coefficient $\lambda$.
Following the practice in~\cite{liu2017learning}, an $L1$ regularization is imposed
on the scaling factors of the batch normalization layers.
As shown in Alg.~\ref{alg:alg-1}, at the end of the $t^{th}$ epoch,
we calculate the overall sparsity of the model:
\begin{equation}
  P = \frac{\sum_l |\mathcal{F}_{\text{pruned}}^l|}{\sum_l C^l}.
  \label{eq:model-sparsity}
\end{equation}
Given the total number of epochs $N$, we compute the expected sparsity gain,
and if the sparsity gain within an epoch does not meet the requirement,
\ie~$P_t - P_{t-1} < (r-P_{t-1}) / (N-t+1)$,
the regularization coefficient $\lambda$ is increased by $\Delta_{\lambda}$.
If the model is over-sparse, \ie~$P_t > r$,
the coefficient $\lambda$ is decreased by $\Delta_{\lambda}$.
\begin{figure}[!t]
  \removelatexerror
  \begin{algorithm}[H]
    \setstretch{1.3}
    \caption{Automatic Regularization Control}\label{alg:alg-1}
    Initialize $\lambda_1 = 0$, $P_1=0$, $N=\text{\#epochs}$\\
    \For( \textit{train for 1 epoch}){$t := 1$ to $N$}
    {
      $P_t = \frac{\sum_l |\mathcal{F}_{\text{pruned}}^l|}{\sum_l C^l}$ \\
      \uIf{$P_t - P_{t-1} < \frac{r-P_{t-1}}{N-t+1}$}{
        $\lambda_{t+1} = \lambda_t + \Delta_{\lambda}$\
      }
      \uElseIf{$P_t > r$}{
        $\lambda_{t+1} = \lambda_t - \Delta_{\lambda}$\
      }
    }
  \end{algorithm}
\end{figure}
This strategy guarantees that the model meets the desired model sparsity,
and that the pruned filters contribute negligibly to the outputs.
%--------------------------------------------------------------------------------------------------------------------%
%--------------------------------------------------------------------------------------------------------------------%
\Uspace
\section{Experimental Results}
\Lspace

In this section, we first describe the details of our implementation
in \secref{sec:details},
and report the experimental results on the CIFAR \cite{krizhevsky2009learning} datasets in \secref{sec:cifar}
and the ImageNet \cite{ILSVRC15} dataset in \secref{sec:imagenet}.

\begin{figure*}[!tb]
  \centering
  \begin{overpic}[width=1\linewidth]{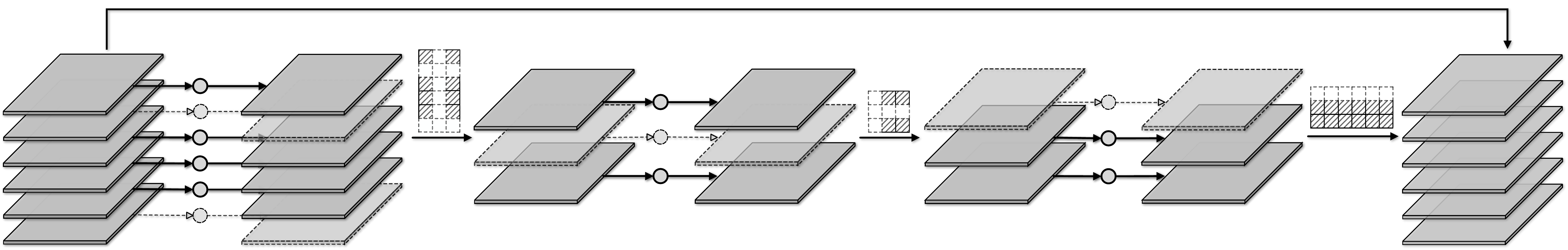}
    \put(2,-1.5){\footnotesize{Input}}
    \put(92,-1.5){\footnotesize{Output}}
    \put(45,13.6){\footnotesize{residual path}}
    \put(11,0){\footnotesize{BN1}}
    \put(8.5,13){\footnotesize{feature selection}}
    %\put(7,-2){\footnotesize{(feature selection)}}
    \put(26,5){\footnotesize{Conv1}}
    \put(40.5,0.5){\footnotesize{BN2}}
    \put(54.2,5){\footnotesize{Conv2}}
    \put(69,0.5){\footnotesize{BN3}}
    \put(84,5){\footnotesize{Conv3}}
  \end{overpic}
  \vspace{1pt}
  \caption{
    Illustration of pruning the bottleneck structure.
    Planes and grids represent feature maps and convolutional kernels, respectively.
    The dotted planes and blank grids denote the pruned feature channels and
    the corresponding convolutional filters.
    We perform ``feature selection'' after the first batch-norm layer,
    and prune only the input dimension of the last convolutional layer.
    Consequently, the number of channels is unchanged in the residual path.
  }\label{fig:residual-prune}
\end{figure*}

%--------------------------------------------------------------------------------------------------------------------%
\Uspace
\subsection{Implementation Details}\label{sec:details}
\Lspace

Our implementation is based on the official training sources of Network Slimming
in the PyTorch \cite{steiner2019pytorch} library.\footnote{\url{https://github.com/Eric-mingjie/rethinking-network-pruning}}
We follow the ``train, prune, and finetune" pipeline as depicted in~\figref{fig:pipeline}.

\PARAspace
\paragraph{Datasets and Data Augmentation}
We conduct image classification experiments on the CIFAR \cite{krizhevsky2009learning},
SVHN \cite{netzer2011reading}, and ImageNet \cite{ILSVRC15} datasets.
For the CIFAR and SVHN datasets, we follow the common practice of data augmentation:
zero-padding of 4 pixels on each side of the image and random cropp of a $32 \times 32$ patch.
On the ImageNet dataset, we adopt the standard data augmentation strategy as in
the prior work~\cite{he2015delving,he2016deep,he2016identity,simonyan2014very}:
resize images to have the shortest edge of 256 pixels and then randomly crop a $224 \times 224$ patch.
Besides, we adopt random horizontal flip on the cropped image for the CIFAR and ImageNet datasets.
The input data is normalized by subtracting the channel-wise means and dividing the channel-wise standard
deviations before being fed to the network.

\PARAspace
\paragraph{Backbone Architectures}
We evaluate the proposed method on two representative architectures:
VGGNet~\cite{simonyan2014very} and ResNet~\cite{he2016deep}.
Following the practice of Network Slimming~\cite{liu2017learning},
we use the Pre-Act-ResNet architecture~\cite{he2016identity} in which
the BN layers and non-linearities are placed before the convolutional layers.
(See \figref{fig:residual-prune}.)

\PARAspace
\paragraph{Hyper-parameters}
The threshold in~\eqref{eq:filters-pruned} is set to $0.01$ unless
otherwise specified, and $\Delta_{\lambda} = 10^{-5}$ in all experiments.
We use the SGD optimizer with a momentum of $0.9$ and a weight decay of $10^{-4}$.
The initial learning rate is $0.1$ and divided by a factor of $10$
at the specified epochs.
We train for $160$ epochs on the CIFAR datasets and $40$ epochs on the SVHN dataset.
The learning rate decays at $50\%$ and $75\%$ of the total training epochs.
On the ImageNet dataset, we train for $100$ epochs and decay the learning rate
every $30$ epochs.

\PARAspace
\paragraph{Half-precision Training on ImageNet}
We train models on the ImageNet dataset with half-precision (FP16),
using the Apex library,\footnote{\url{https://github.com/NVIDIA/apex}}
where parameters of batch normalization are represented in FP32
while others in FP16.
This allows us to train the ResNet-50 model within $40$ hours 
on $4$ RTX $2080$Ti GPUs.
Despite training with FP16, we do not observe obvious performance
degradation in our experiments.
For example, as shown in~\tabref{tab:imagenet}, we achieve a top-1 accuracy of
$76.27\%$ with the Pre-ResNet-50 architecture on the ImageNet dataset,
which is very close to that in the original paper~\cite{he2016identity}
or reported in~\cite{molchanov2019importance}.

\PARAspace
\paragraph{Train, Prune, and, Finetune}
We adopt the three-stage pipeline, \ie~train, prune, and finetune,
as in many previous pruning methods
~\cite{he2018soft,he2019asymptotic,he2019filter,liu2017learning,luo2017thinet,zeng2018accelerating}.
(See \figref{fig:pipeline}.)
% We follow the common practice of three-stage pipeline (\figref{fig:pipeline}) in many previous pruning
% methods~\cite{he2018soft,he2019filter,liu2017learning,luo2017thinet}.
%
In the experiments, we found that in the first stage, the model sparsity $P$
grows rapidly when the learning rate is large.
After the learning rate decays, the model sparsity hardly increases
unless an extremely large $\lambda$ is reached.
Therefore, to effectively promote model sparsity,
we keep the learning rate fixed in the first stage,
and decays the learning rate normally when in the third stage.
On CIFAR datasets, we train for 160 epochs for the first stage,
and on the ImageNet dataset, we train only 40 epochs for the first stage.
On both CIFAR and ImageNet datasets, we finetune for a full episode.

\begin{table}[thpb]
  \renewcommand{\arraystretch}{1.2}
   \centering
   \caption{
     Experimental results on the CIFAR10 dataset.
     Our method performs favorably against the Network Slimming (SLM)~\cite{liu2017learning}
     baseline and other pruning methods on both VGGNets and ResNets.
   }
   \resizebox{0.48\textwidth}{!}{
   \begin{tabular}{l|c|c|cc}
   \hline
   % \rowcolor[rgb]{ .85,  .9,  0.95}
    & & & \textbf{Baseline} & \textbf{Finetune} \\
   %  \rowcolor[rgb]{ .85,  .9,  0.95}
    \multirow{-2}{*}{\textbf{Model}} & \multirow{-2}{*}{\textbf{Methods}} & \multirow{-2}{*}{\textbf{ratio $r$}} & \textbf{accuracy} & \textbf{accuracy} \\
   
    \hline
   \multirow{2}{*}{VGG11}
   & SLM~\cite{liu2017learning} & \multirow{2}{*}{0.5} & 92.13 ($\pm0.18$)  & 91.91 ($\pm0.01$) \\
   
   & Ours &  & 92.02 ($\pm0.20$) & 92.17 ($\pm0.18$) \\
   \hline
   \multirow{2}{*}{VGG16}
   & SLM & \multirow{2}{*}{0.6} & 93.73 ($\pm0.06$) & 93.65 ($\pm0.04$) \\
   & Ours & & 93.57 ($\pm0.26$) & 93.70 ($\pm0.05$) \\
   \hline
   \multirow{2}{*}{VGG19}
   & SLM (\footnotesize{from~\cite{liu2018rethinking}}) & \multirow{2}{*}{0.7}& 93.53  ($\pm0.16$) & 93.60 ($\pm0.16$) \\
   & Ours &  & 93.66 ($\pm0.26$) & 93.53 ($\pm0.24$) \\
   \hline
   \multirow{8}{*}{Res56}
   % cifar10 baseline copy from geometric median.
   & SFP & \multirow{5}{*}{0.4} & 93.59  ($\pm0.58$) & 92.26 ($\pm0.31$) \\
   & ASFP~\cite{he2019asymptotic} &  & 93.59  ($\pm0.58$) & 92.44 ($\pm0.31$) \\
   & FPGM~\cite{he2019filter} & & 93.59  ($\pm0.58$) & 92.93 ($\pm0.49$) \\
   & SLM &  & 93.56 ($\pm0.19$) & 93.33 ($\pm0.14$) \\
   & Ours & & 93.73 ($\pm0.10$) & 93.86 ($\pm0.19$) \\
   \cdashline{2-5}[1pt/2pt]
   & SLM & \multirow{2}{*}{0.5} & 93.56  ($\pm0.19$) & 92.90 ($\pm0.14$) \\
   & Ours & & 93.73 ($\pm0.10$) & 93.62 ($\pm0.16$) \\
   \cdashline{2-5}[1pt/2pt]
   & SLM & \multirow{2}{*}{0.6} & 93.56  ($\pm0.19$) & 91.94 ($\pm0.10$) \\
   & Ours & & 93.73 ($\pm0.10$) & 92.68 ($\pm0.15$) \\
   \hline
   \multirow{8}{*}{Res110}
   % cifar10 baseline copy from geometric median.
   & SFP & \multirow{5}{*}{0.4} & 93.68  ($\pm0.32$) & 93.38 ($\pm0.30$) \\
   & ASFP~\cite{he2019asymptotic} &  & 93.68  ($\pm0.58$) & 93.20 ($\pm0.10$) \\
   & FPGM~\cite{he2019filter} & & 93.68 ($\pm0.32$) & 93.73 ($\pm0.23$) \\
   & SLM &  & 94.61 ($\pm0.01$) & 94.49 ($\pm0.12$) \\
   & Ours & & 94.43 ($\pm0.13$) & 94.75 ($\pm0.12$) \\
   \cdashline{2-5}[1pt/2pt]
   & SLM & \multirow{2}{*}{0.5} & 94.61 ($\pm0.01$) & 94.24 ($\pm0.13$) \\
   & Ours & & 94.43 ($\pm0.13$) & 94.52 ($\pm0.26$) \\
   \cdashline{2-5}[1pt/2pt]
   & SLM & \multirow{2}{*}{0.6} & 94.61 ($\pm0.01$) & 93.47 ($\pm0.15$) \\
   & Ours & & 94.43 ($\pm0.13$) & 94.57 ($\pm0.04$) \\
   \hline
   \multirow{6}{*}{Res164}
   & SLM (\footnotesize{from~\cite{liu2018rethinking}}) & \multirow{2}{*}{0.4} & 95.04 ($\pm0.16$) & 94.77 ($\pm0.12$) \\
   & Ours & & 94.86 ($\pm0.10$) & 95.01 ($\pm0.15$) \\
   \cdashline{2-5}[1pt/2pt]
   & SLM & \multirow{2}{*}{0.5} & 95.04 ($\pm0.16$) & 94.52 ($\pm0.09$) \\
   & Ours & & 94.86 ($\pm0.10$) & 94.83 ($\pm0.05$) \\
   \cdashline{2-5}[1pt/2pt]
   & SLM (\footnotesize{from~\cite{liu2018rethinking}}) & \multirow{2}{*}{0.6} & 95.04 ($\pm0.16$) & 94.23 ($\pm0.21$)\\
   & Ours & & 94.86 ($\pm0.10$) & 94.53 ($\pm0.35$) \\
   \hline
   %-----------------------------------------------%
   \end{tabular}}
   \label{tab:cifar10}
\end{table}

\PARAspace
\paragraph{Prune with Short Connections}
In the Pre-Act-ResNet architecture, operators are arranged in the
``BN, ReLU, and Conv'' order.
As depicted in \figref{fig:residual-prune}, given the input feature maps,
we perform a ``feature selection'' right after the first batch normalization layer
(BN1) to filter out less important channels according to the
dependency-aware channel importance (\eqref{eq:channel-importance}).
For the first and second convolutional layers (Conv1 and Conv2),
we prune both the input and output dimensions of their kernels.
(The pruned channels are represented as the dotted planes in \figref{fig:residual-prune}.)
For the last convolutional layer (Conv3), we prune only the input dimension
of Conv3 to preserve the structure of the residual path.
After pruning, the number of channels in the residual path remains unchanged.
Note that when computing the model sparsity (\eqref{eq:model-sparsity}), the ``feature selection''
is not taken into account because it does not actually prune any filters.
For example, in the case of \figref{fig:residual-prune},
there are only 2 filters pruned, \ie~the second filter of Conv2 and the first filter of Conv3.

%--------------------------------------------------------------------------------------------------------------------%
\Uspace
\subsection{Results on CIFAR}\label{sec:cifar}
\Lspace

\begin{figure}[!t]
  \centering
  \begin{overpic}[width=1\linewidth]{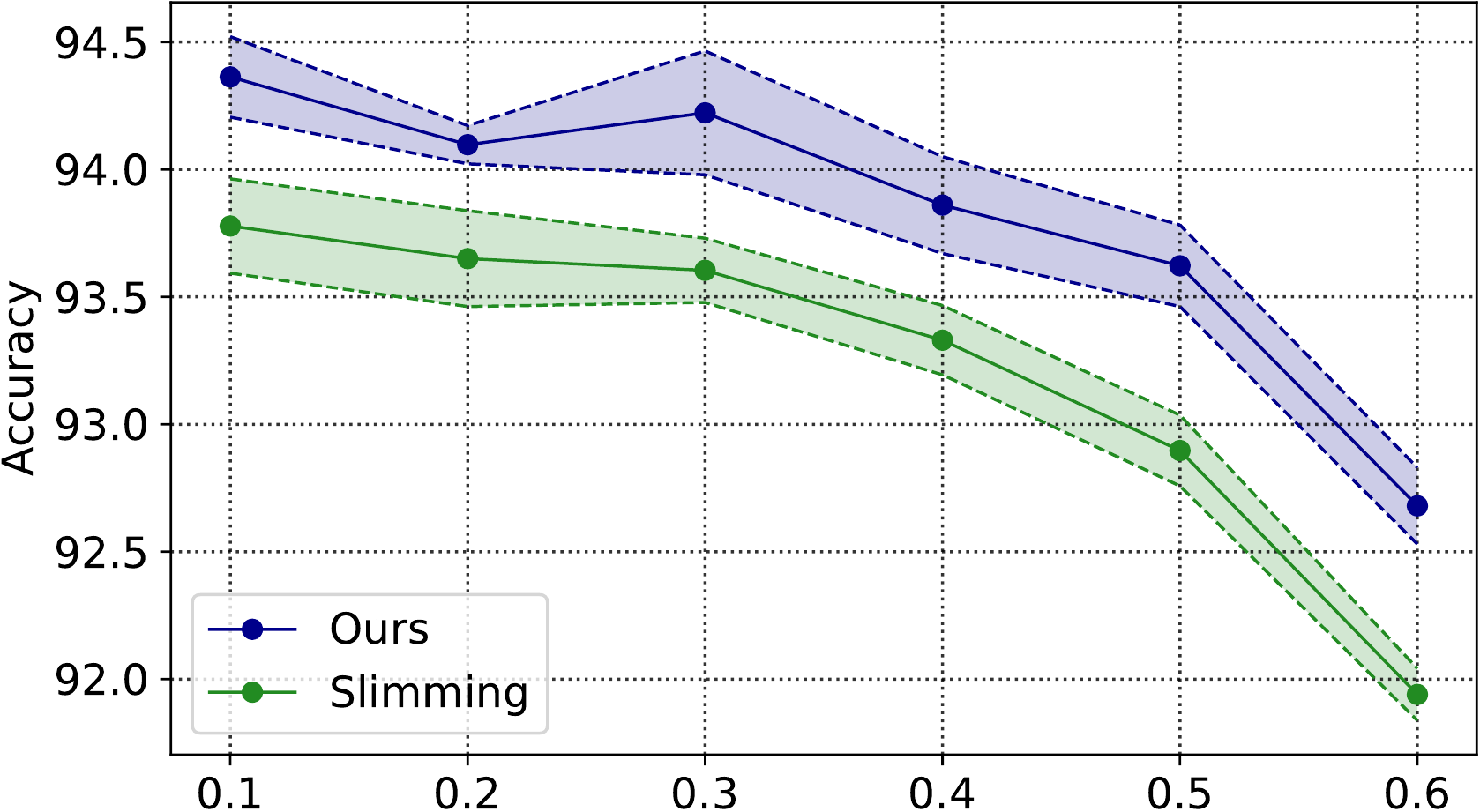}
  \put(45,-4){pruning ratio $r$}
  \end{overpic}
  \vspace{2pt}
  \caption{
  Performance (mean and standard deviation over a 10-fold validation) of pruning
  the ResNet-56 network on the CIFAR10 dataset under various pruning ratios $r$.
  }\label{fig:sensitivity}
\end{figure}

We first evaluate our method on the CIFAR10 and CIFAR100 datasets.
Experiments on the CIFAR datasets are conducted using the VGGNets and ResNets with various
depths.
On the CIFAR datasets, we record the mean and standard deviation over a 10-fold validation.
It is worthy of noting that, as described in~\secref{sec:filter-selection},
Network Slimming \cite{liu2017learning} often results in unstable architectures,
whose performance is greatly degraded.
(See \secref{sec:prune-stable} for details.)
Therefore, for Network Slimming, we skip the outliers and restart the pipeline
if the accuracy is $10\%$ lower than the mean accuracy.
Quantitative results on CIFAR10 and CIFAR100 datasets are
summarized in~\tabref{tab:cifar10} and~\tabref{tab:cifar100}, respectively.
Additionally, a curve of the classification accuracy \textit{v.s.} the
pruning ratio $r$ is shown in~\figref{fig:sensitivity}.

\begin{table}[!t]
  \renewcommand{\arraystretch}{1.3}
   \centering
   \caption{
     Experimental results on the CIFAR100 dataset.
     Here, ``N/A'' indicates the compressed model collapses in all runs.
     Still, our approach consistently outperforms the Network Slimming (SLM)~\cite{liu2017learning}
     baseline.
     Notably, our approach outperforms Network Slimming by up to $2\%$ on the ResNet-164 backbone.
   }
   \resizebox{0.48\textwidth}{!}{
   \begin{tabular}{l|c|c|cc}
   \hline
   % \rowcolor[rgb]{ .85,  .9,  0.95}
   & & & \textbf{Baseline} & \textbf{Finetune} \\
   %  \rowcolor[rgb]{ .85,  .9,  0.95}
    \multirow{-2}{*}{\textbf{Model}} & \multirow{-2}{*}{\textbf{Methods}} & \multirow{-2}{*}{\textbf{ratio $r$}} & \textbf{accuracy} & \textbf{accuracy} \\
     
    \hline
   \multirow{2}{*}{VGG11}
   & SLM \cite{liu2017learning} & \multirow{2}{*}{0.3} & 69.33 ($\pm0.26$)  & 66.54 ($\pm0.14$) \\
   & Ours &  & 68.24 ($\pm0.11$) & 67.84 ($\pm0.11$) \\
   \hline
   \multirow{4}{*}{VGG16}
   & SLM & \multirow{2}{*}{0.3} & 73.50 ($\pm0.18$) & 73.36 ($\pm0.28$) \\
   & Ours & & 72.16 ($\pm0.23$) & 73.59 ($\pm0.37$) \\
   \cdashline{2-5}[1pt/2pt]
   & SLM & \multirow{2}{*}{0.4} & 73.50 ($\pm0.18$) & N/A \\
   & Ours & & 72.16 ($\pm0.23$) & 73.59 ($\pm0.23$) \\
   \hline
   \multirow{2}{*}{VGG19}
   & SLM (\footnotesize{from~\cite{liu2018rethinking}}) & \multirow{2}{*}{0.5}& 72.63 ($\pm0.21$) & 72.32 ($\pm0.28$) \\
   & Ours &  & 71.19 ($\pm0.54$) & 72.48 ($\pm0.28$) \\
   \hline
   \multirow{4}{*}{Res164}
   & SLM (\footnotesize{from~\cite{liu2018rethinking}}) & \multirow{2}{*}{0.4} & 76.80 ($\pm0.19$) & 76.22 ($\pm0.20$) \\
   & Ours & & 76.43 ($\pm0.26$) & 77.74 ($\pm0.17$) \\
   \cdashline{2-5}[1pt/2pt]
   & SLM & \multirow{2}{*}{0.6} & 76.80 ($\pm0.19$) &  74.17 ($\pm0.33$)\\
   & Ours & & 76.43 ($\pm0.26$) & 76.28 ($\pm0.27$) \\
   \hline
   %-----------------------------------------------%
   \end{tabular}}
   \label{tab:cifar100}
\end{table}

\PARAspace
\paragraph{VGGNets}
We start with the simpler architecture, VGGNet, which is a sequential
architecture without skip connections.
We find that pruning a large number of filters brings a puny performance drop.
Take the VGGNet-19 as an example.
On the CIFAR10 dataset, with 70\% of the filters pruned,
both Network Slimming and our method even bring a little performance gain.
And interestingly, increasing model depth does not always enhance performance.
On both CIFAR10 and CIFAR100 datasets, VGGNet-16 achieves better (or comparable)
performance than VGGNet-19.
These observations demonstrate the VGGNet is heavily over-parameterized for the CIFAR datasets,
and that pruning a proportion of filters brings negligible influence to the performance.

\PARAspace
\paragraph{ResNets}
Pruning the ResNet architectures is more complicated because of the residual paths.
As described in~\secref{sec:details} and ~\figref{fig:residual-prune}, we preserve the
number of channels in the residual path and only prune filters inside the bottleneck architecture.
By pruning the same proportion of filters, our method consistently achieves better results
compared with the Network Slimming \cite{liu2017learning} baseline.

%---------------------------------------------------------------------------%
\Uspace
\subsection{Results on SVHN} \label{sec:svhn}
\Lspace

We then apply the proposed pruning algorithm to the ResNet family on the SVHN dataset,
following the same evaluation protocol as in \secref{sec:cifar}.
It can be seen from \tabref{tab:svhn} that our approach outperforms 
the state-of-the-art baseline method \cite{liu2017learning}
under various model depths and pruning ratios.
Also, Network Slimming \cite{liu2017learning} often collapses when the pruning ratio
is high, \eg~$80\%$, while our approach is more tolerant of high pruning ratios
and still maintains a competitive accuracy.
For example, only an accuracy of $0.10\%$ is sacrisficed for $80\%$ of filters
being pruned from the ResNet-56 backbone.
Furthermore, similar to the circumstances on the CIFAR datasets,
pruning a proportion of filters may even bring a performance gain
(\eg~when $20\%$ or $40\%$ of filters are pruned),
indicating a moderate pruning ratio can alleviate the over-fitting problem
on the relatively small datasets, such as CIFAR and SVHN.

\begin{table}[!t]
  \renewcommand{\arraystretch}{1.2}
   \centering
   \caption{
     Experimental results on the SVHN dataset.
     Similarly, ``N/A'' indicates the compressed model collapses in all runs.
     It can be seen that our approach is tolerant of high pruning ratios and
     outperforms the Netwoek Slimming (SLM) \cite{liu2017learning} baseline
     under various experimental settings.
   }
   \resizebox{0.48\textwidth}{!}{
   \begin{tabular}{l|c|c|cc}
   \hline
   % \rowcolor[rgb]{ .85,  .9,  0.95}
    & & & \textbf{Baseline} & \textbf{Finetune} \\
   %  \rowcolor[rgb]{ .85,  .9,  0.95}
    \multirow{-2}{*}{\textbf{Model}} & \multirow{-2}{*}{\textbf{Methods}} & \multirow{-2}{*}{\textbf{ratio $r$}} & \textbf{accuracy} & \textbf{accuracy} \\
   
    \hline
   \multirow{8}{*}{Res20}
   & SLM & \multirow{2}{*}{0.2} & 95.85 ($\pm0.07$) & 95.82 ($\pm0.18$) \\
   & Ours & & 95.85 ($\pm0.07$) & 96.18 ($\pm0.09$) \\
   \cdashline{2-5}[1pt/2pt]
   & SLM & \multirow{2}{*}{0.4} & 95.85 ($\pm0.07$) & 95.77 ($\pm0.13$) \\
   & Ours & & 95.85 ($\pm0.07$) & 96.20 ($\pm0.11$) \\
   \cdashline{2-5}[1pt/2pt]
   & SLM & \multirow{2}{*}{0.6} & 95.85 ($\pm0.07$) & 95.66 ($\pm0.07$) \\
   & Ours & & 95.85 ($\pm0.07$) & 96.15 ($\pm0.05$) \\
   \cdashline{2-5}[1pt/2pt]
   & SLM & \multirow{2}{*}{0.8} & 95.85 ($\pm0.07$) & N/A \\
   & Ours & & 95.85 ($\pm0.07$) & 95.49 ($\pm0.13$) \\
   \hline
   \multirow{8}{*}{Res56}
   & SLM & \multirow{2}{*}{0.2} & 96.87 ($\pm0.04$) & 96.62 ($\pm0.05$) \\
   & Ours & & 96.87 ($\pm0.04$) & 97.04 ($\pm0.08$) \\
   \cdashline{2-5}[1pt/2pt]
   & SLM & \multirow{2}{*}{0.4} & 96.87 ($\pm0.04$) & 96.56 ($\pm0.07$) \\
   & Ours & & 96.87 ($\pm0.04$) & 97.00 ($\pm0.02$) \\
   \cdashline{2-5}[1pt/2pt]
   & SLM & \multirow{2}{*}{0.6} & 96.87 ($\pm0.04$) & N/A \\
   & Ours & & 96.87 ($\pm0.04$) & 97.03 ($\pm0.02$) \\
   \cdashline{2-5}[1pt/2pt]
   & SLM & \multirow{2}{*}{0.8} & 96.87 ($\pm0.04$) & N/A \\
   & Ours & & 96.87 ($\pm0.04$) & 96.77 ($\pm0.05$) \\
   \hline
   %-----------------------------------------------%
   \end{tabular}}
   \label{tab:svhn}
\end{table}

%--------------------------------------------------------------------------------------------------------------------%
\Uspace
\subsection{Results on ImageNet}\label{sec:imagenet}
\Lspace

Here, we evaluate the proposed method on the large-scale and challenging ImageNet~\cite{ILSVRC15}
benchmark.
The results of Network Slimming \cite{liu2017learning} and our method are obtained
from our implementation, while other results come from the original papers.
We compare against several recently-proposed pruning methods with various criterion,
including the weight norm~\cite{li2017pruning},
norm of batch-norm factors~\cite{liu2017learning,ye2018rethinking},
and a data-dependent pruning method~\cite{molchanov2019importance}.
As summarized in~\tabref{tab:imagenet}, under the same pruning
ratios, our method consistently outperforms the Network Slimming baseline,
and retains a comparable number of parameters and complexity (FLOPs).
Even compared with the data-dependent pruning method~\cite{molchanov2019importance},
our method still achieves competitive performance.
 
\begin{table}[!t]
  \renewcommand{\arraystretch}{1.3}
  \centering
  \caption{
    Image classification results on the ImageNet dataset.
    Our method consistently outperforms the data-independent pruning methods~\cite{luo2017thinet,ye2018rethinking,liu2017learning,li2017pruning},
    and achieves competitive performance against the data-dependent method~\cite{molchanov2019importance}.
  }
  \newcommand{\CC}[1]{\cellcolor{gray!#1}}
  \resizebox{0.475\textwidth}{!}{
  \begin{tabular}{l|ccccc}
  \hline
  \textbf{Model} & \textbf{Methods} & \textbf{ratio} $r$ & \textbf{Acc.} (\%) & \makecell{\textbf{\#Params}\\($10^7$)} & \makecell{\textbf{\#FLOPs}\\($10^6$)}\\
  \hline
  \multirow{3}{*}{VGG11}
  & \CC{30}Baseline & \CC{30}- & \CC{30}70.84 & \CC{30}3.18 & \CC{30}7.61 \\
  & SLM~\cite{liu2017learning} & 0.50 & 68.62 & 1.18 & 6.93 \\
  & Ours & 0.50 & 69.12 & 1.18 & 6.97 \\
  \hline
  \multirow{11}{*}{Res50}
  & \CC{30}Baseline & \CC{30}- & \CC{30}76.27 & \CC{30}2.56 & \CC{30}4.13 \\
  & ThiNet~\cite{luo2017thinet} & 0.50 & 71.01 & 1.24 & 3.48 \\
  & ThiNet & 0.70 & 68.42 & 0.87 & 2.20 \\
  & Li~\etal~\cite{li2017pruning} & N/A & 72.04 & 1.93 & 2.76 \\
  & SSR-L2,1 ~\cite{lin2019toward} & N/A & 72.13 & 1.59 & 1.9 \\
  & SSR-L2,0 ~\cite{lin2019toward} & N/A & 72.29 & 1.55 & 1.9 \\
  & SLM & 0.50 & 71.99 & 1.11 & 1.87 \\
  & Ours & 0.50 & 72.41 & 1.07 & 1.86 \\
  \cdashline{2-6}[1pt/2pt]
  & Taylor~\cite{molchanov2019importance} & 0.19 & 75.48 & 1.79 & 2.66 \\
  & SLM & 0.20 & 75.12 & 1.78 & 2.81 \\
  %& DA & 0.2 & 75.46 & 1.73 & 2.65 \\
  & Ours & 0.20 & 75.37 & 1.76 & 2.82 \\
  \hline
  \multirow{9}{*}{Res101}
  & \CC{30}Baseline & \CC{30}- & \CC{30}77.37 & \CC{30}4.45 & \CC{30}7.86 \\
  & Ye~\etal~\cite{ye2018rethinking}-v1 & N/A & 74.56 & 1.73 & 3.69 \\  
  & Ye~\etal~\cite{ye2018rethinking}-v2 & N/A & 75.27 & 2.36 & 4.47 \\
  & Taylor~\cite{molchanov2019importance} & 0.45 & 75.95 & 2.07 & 2.85 \\
  & SLM & 0.50 & 75.97 & 2.09 & 3.16 \\
  & Ours & 0.50 & 76.54 & 2.17 & 3.23 \\
  \cdashline{2-6}[1pt/2pt]
  & Taylor~\cite{molchanov2019importance} & 0.25 & 77.35 & 3.12 & 4.70 \\
  & Ours & 0.20 & 77.36 & 3.18 & 4.81 \\
  \hline
  %-----------------------------------------------%
  \end{tabular}}
  \label{tab:imagenet}
\end{table}

%--------------------------------------------------------------------------------------------------------------------%
%--------------------------------------------------------------------------------------------------------------------%
\Uspace
\section{Ablation Study}
\Lspace

In this section, we conduct several ablation studies to justify our design choice.
All the experiments in this section are conducted on the CIFAR100 dataset.

\begin{figure*}[!t]
  \centering
  \begin{overpic}[width=1\linewidth]{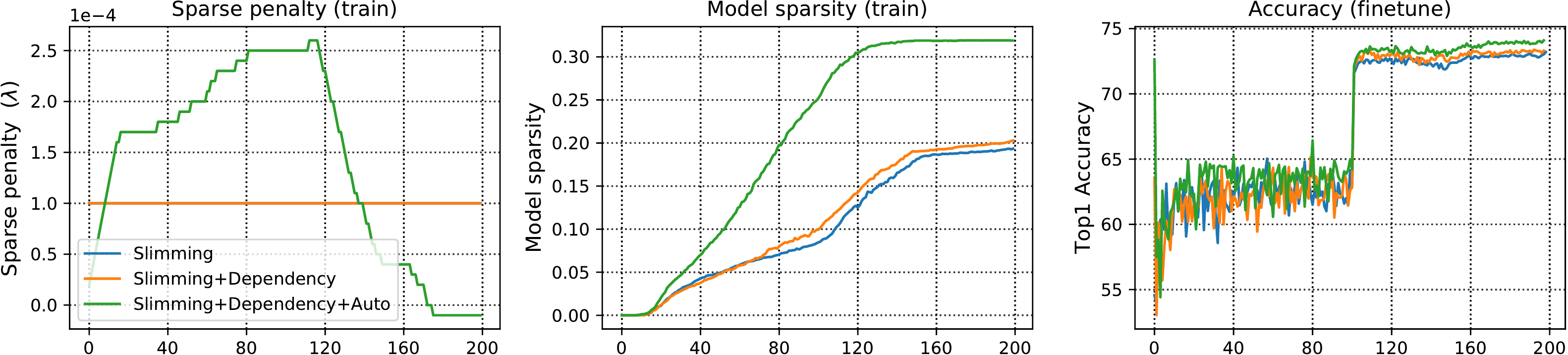}
  \put(7,-2.5){(a) Sparsity Regularization}
  \put(45,-2.5){(b) Model Sparsity}
  \put(77.5,-2.5){(c) Finetune Accuracy}
  \end{overpic}
  \vspace{3pt}
  \caption{
    Training dynamics of pruning the VGGNet-16 backbone ($r=0.3$) on the CIFAR100 dataset
    with the three different strategies.
    The horizontal axis represents the training epochs in all three plots.
    Plot (a), (b), and (c) represent the regularization coefficient $\lambda$,
    model sparsity $P$, and the finetune accuracy, respectively.
    Compared with the Network Slimming baseline, the dependency-aware
    importance estimation assists to identify less important filters,
    leading to higher performance before/after finetuning.
    Then, equipped with the automatic regularization control,
    the model meets the desired sparsity at the end of the first stage,
    and achieves the best performance after finetuning.
  }
  \label{fig:curve1}
\end{figure*}

\Uspace
\subsection{The Effectiveness of Dependency-aware Importance Estimation}\label{sec:importance-estimation}
\Lspace

In the first ablation study, we verify that our method can more accurately identify
less important filters, thus leading to a better compressed architecture.
This can be evidenced by 1) the less performance drop after pruning,
and 2) the better final performance after finetuning.

With the same pruning ratio, \eg~$r=0.5$, we assume that the importance estimation is more accurate
if the pruned model (without finetuning) achieves higher performance on the validation set.
Thus, the accuracy of importance estimation can be measured by the performance of pruned
networks under the same pruning ratio.
In this experiment, we compare the following three strategies:
(a) Network Slimming~\cite{liu2017learning} which measures filter importance
by the batch-norm scaling factors only;
(b) the dependency-aware importance estimation in \eqref{eq:channel-importance}; and
(c) the dependency-aware importance estimation + automatic regularization control.

Firstly, we conduct an illustrative experiment on the VGGNet-16 backbone with a pruning ratio of $0.3$.
As shown in~\figref{fig:curve1}, the strategy (c) obtains a compressed model
with the desired sparsity and achieves the best accuracy after finetuning.
Then, we quantitatively compare these three strategies on the VGGNet-16 and ResNet-56 backbones.
The statistics over a 10-fold validation are reported in~\tabref{tab:ablation-1}.

\begin{table}[!t]
  \centering
  \caption{
    The performance of different strategies before and after finetuning
    are demonstrated in the table.
  }
  \renewcommand{\arraystretch}{1.15}
  \resizebox{0.48\textwidth}{!}{
  \begin{tabular}{l|cccccc}
  \hline
  \textbf{Model} & \textbf{Methods} & \textbf{ratio} $r$  &
  \makecell{\textbf{Before}\\\textbf{Finetune}} &
  \makecell{\textbf{After}\\\textbf{Finetune}}\\
  \hline
  \multirow{3}{*}{VGG16}
  & SLM & 0.3 & 52.19 ($\pm6.82$) & 73.36 ($\pm0.28$)\\
  \cdashline{2-5}[1pt/2pt]
  & \makecell{SLM+DA} & 0.3  & 61.19 ($\pm6.18$) & 73.57 ($\pm0.31$)\\
  \cdashline{2-5}[1pt/2pt]
  & \makecell{SLM+DA+Auto}& 0.3  & 72.83 ($\pm0.26$) & 73.59 ($\pm0.37$) \\
  \hline
  \multirow{3}{*}{Res56}
  & SLM & 0.5 & 1.41 ($\pm0.25$) & 71.13 ($\pm0.26$)\\
  \cdashline{2-5}[1pt/2pt]
  & \makecell{SLM+DA} & 0.5 & 5.29 ($\pm1.01$) & 73.62 ($\pm0.14$)\\
  \cdashline{2-5}[1pt/2pt]
  & \makecell{SLM+DA+Auto} & 0.5 & 55.29 ($\pm1.92$) & 74.53 ($\pm0.10$)\\
  \hline
  %-----------------------------------------------%
  \end{tabular}}
  \label{tab:ablation-1}
\end{table}

The results in~\tabref{tab:ablation-1} reveal that
1) the dependency-aware importance estimation is able to measure
the filter importance more accurately as it achieves a much higher performance before finetuning
compared with the Network Slimming, and
2) the automatic regularization control assists to derive a model with desired sparsity
and search for a better architecture, evidenced by the favorable performance after finetuning.
% We compare the pretrained accuracy (accuracy after first round training),
% pruned accuracy (accuracy after removing less important filters)
% and finetune accuracy (the final accuracy after finetuning).
% %
% The quantitative results are in ~\tabref{tab:ablation-1}.
%
% Due to the automatic sparsity control, our method may 
% introduce larger $L_1$ coefficient ($\lambda$)
% during training, hence the pretrain accuracy is a little bit lower than the baseline.
%
% However, after pruning, our method (method-c) presents significant higher accuracy.

\Uspace
\subsection{Fixed \textit{v.s.} Adjustable Regularization Coefficient}
\Lspace

There are two alternative approaches that can help achieve the desired mode sparsity:
(a) fix the threshold $p$ and adjust the regularization coefficient $\lambda$ during training;
and (b) fix $\lambda$ and search for a suitable $p$ after training.

We compare these two alternatives on the ResNet-56 backbone with a pruning ratio of $0.5$,
which means $50\%$ of the filters will be pruned.
For strategy (a), the regularization coefficient $\lambda$ is fixed to $10^{-5}$,
as suggested by~\cite{liu2017learning}.

\begin{table}[!b]
  \centering
  \caption{
    Comparison of the two alternatives of reaching the desired model sparsity.
    % A: set a fixed $p$ and adjust penalty $\lambda$ during training.
    % B: train with a fixed $\lambda$ and search a threshold $p$ to obtain desired sparsity.
  }
  \renewcommand{\arraystretch}{1.2}
  \resizebox{0.38\textwidth}{!}{
  \begin{tabular}{l|cccc}
  \hline
  \textbf{Method}  & \makecell{\textbf{Before}\\\textbf{Pruning}}& \textbf{threshold} $p$ &
  \makecell{\textbf{Before}\\\textbf{Finetune}} & \makecell{\textbf{After}\\\textbf{Finetune}}\\
  \hline
  (a) & 60.86 & 0.01 & 60.86  & 75.24\\
  (b) & 73.59 & 0.41 &1.53  & 74.36\\
  \hline
  %-----------------------------------------------%
  \end{tabular}}
  \label{tab:ablation-ab}
\end{table}

As shown in~\tabref{tab:ablation-ab}, under the same pruning ratio,
strategy (a) performs favorably against strategy (b) in terms of
the performance before and after finetuning.
% strategy (a) holds significantly higher accuracy than that of method-B
% and achieves better performance after finetuning.
%
This justifies our design of dynamically adjusting $\lambda$ during training.

\Uspace
\subsection{Pruning as Architecture Search}\label{sec:prune-as-search}
\Lspace

As pointed out in~\secref{sec:filter-selection},
Network Slimming~\cite{liu2017learning} may lead to unreasonable
compressed architectures as too many filters can be pruned in a single layer.
In this experiment, we verify that our method can derive better compressed architectures.
To test the difference of the pruned architectures,
we re-initialize the parameters of pruned models,
and then train the pruned models for a full episode
as in the standard pipeline.
Note that we are essentially training the compressed architecture from scratch
under the ``scratch-E'' setting in \cite{liu2018rethinking}.
% Concretely, after pruning, we randomly re-initialize
% the parameters of pruned architectures and then retrain
% under the ``scratch-E'' setting as in~\cite{liu2018rethinking}.
%
The results in~\tabref{tab:ablation-2} indicate that our method
derives better compressed architectures, as evidenced by the superior
performance when training from scratch.

\begin{table}[!t]
  \centering
  \caption{
    The performance of training the compressed architecture from scratch.
    By training the pruned model with randomly re-initialized weights,
    our method can still outperform the Network Slimming (SLM)~\cite{liu2017learning}
    baseline, implying that our approach derives a better network architecture.
  }
  \footnotesize
  \renewcommand{\arraystretch}{1.25}
  \renewcommand{\tabcolsep}{4pt} 
  \resizebox{0.48\textwidth}{!}{
  \begin{tabular}{l|c|cccc}
  \hline
  \multirow{2}{*}{\textbf{Model}} & \multirow{2}{*}{\textbf{Method}}  &
  \multicolumn{3}{c}{\textbf{Accuracy} (\%)}\\
  % \cline{3-5}
  & & \textbf{Baseline} & \textbf{Finetune} & \textbf{Scratch}\\
  \hline
  \multirow{2}{*}{Res164}
  & SLM & 76.80 ($\pm 0.19$) & 74.17 ($\pm0.33$) & 75.05 ($\pm0.08$)\\  
  \cdashline{2-5}[1pt/2pt]
  & Ours & 76.43 ($\pm 0.26$)  & 76.43 ($\pm0.27$) & 76.41 ($\pm0.32$) \\\hline
  %-----------------------------------------------%
  \end{tabular}}
  \label{tab:ablation-2}
\end{table}

\Uspace
\subsection{Pruning Stability}\label{sec:prune-stable}
\Lspace

As stated in~\secref{sec:filter-selection}, Network Slimming \cite{liu2017learning}
selects filters to be pruned by ranking channel importance of different layers
across the entire network, leading to unstable architectures.
We empirically verify the claim that with a large pruning ratio, our method can still
achieve promising results, while Network Slimming leads to collapsed models
with a high probability.
% fails to converge with a high probability.

\begin{figure}[!b]
  \centering
  \begin{overpic}[width=1\linewidth]{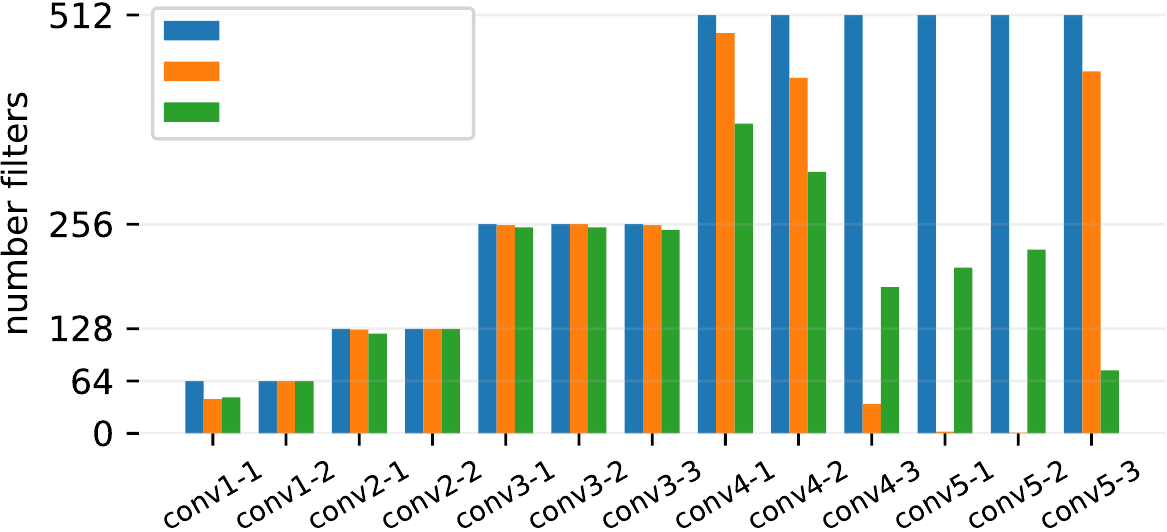}
  \put(20,41.8){\footnotesize{Baseline}}
  \put(20,38.3){\footnotesize{Slimming~\cite{liu2017learning}}}
  \put(20,34.8){\footnotesize{Ours}}
  \end{overpic}
  \caption{
    Filter distributions of the pruned VGGNet-16 backbone.
    Network Slimming \cite{liu2017learning} presents an unbalanced architecture
    where conv5-1 has two filters remained and conv5-2 has only one filter remained.
  }
  \label{fig:pruned-arch}
\end{figure}

\begin{table}[!t]
  \centering
  \caption{
    Record of a 5-fold validation on the CIFAR datasets with the VGGNet-16 backbone.
    In the table, ($\cdot$/$\cdot$) indicates the finrtune accuracy and
    the minimal number of remaining channels in each layer after pruning.
  }
  \renewcommand{\arraystretch}{1.4}
  \renewcommand{\tabcolsep}{4pt} 
  \resizebox{0.48\textwidth}{!}{
  \begin{tabular}{c|c|c|c|c|c|c|c}
  \hline
  \textbf{Dataset} & \textbf{Method} & \textbf{ratio} $r$ & \textbf{run-1} & \textbf{run-2} & \textbf{run-3} & \textbf{run-4} & \textbf{run-5}\\
  \hline
  \multirow{2}{*}{CIFAR10} & SLM & \multirow{2}{*}{0.7} & 10.00~/~0 & 10.00~/~0 & 10.00~/~0 & 10.00~/~0 & 10.00~/~0\\
   & Ours &  & 93.93~/~24 & 93.66~/~25 & 93.94~/~27 & 93.70~/~23 & 93.89~/~27 \\\hline
  \multirow{2}{*}{CIFAR100} & SLM & \multirow{2}{*}{0.4} & 1.00~/~0 & 1.00~/~1 & 1.00~/~0 & 1.00~/~0 & 1.00~/~0\\
   & Ours &  & 73.24~/~29 & 73.60~/~37 & 73.92~/~35 & 73.47~/~37 & 73.71~/~37\\
  \hline
  %-----------------------------------------------%
  \end{tabular}}
  \label{tab:ablation-pruned-arch}
\end{table}

Here, we design two experiments.
In the first experiment, we give an intuitionistic illustration of the compressed
network architecture induced by Network Slimming and our method.
We use the VGGNet-16 backbone with a pruning ratio of $0.4$.
The filter distributions of compressed architectures are shown in~\figref{fig:pruned-arch}.

In the second experiment, we conduct a 5-fold validation on the CIFAR10 and CIFAR100 datasets,
again using the VGGNet-16 backbone.
The results in~\tabref{tab:ablation-pruned-arch} indicate that
under a relatively high pruning ratio, our method can still
achieve high performance while Network Slimming collapses in all runs.

\Uspace
\section{Conclusion}
\Lspace

In this paper, we propose a principled criteria to identify
the unimportant filters with consideration of the inter-layer dependency.
Based on this, we prune filters based on the local channel importance,
and introduce an automatic-regularization-control mechanism to dynamically
adjust the coefficient of sparsity regularization.
In the end, our method is able to compress the state-of-the-art neural networks
with a minimal accuracy drop.
Comprehensive experimental results on CIFAR, SVHN, and ImageNet datasets
demonstrate that our approach performs favorably against the Network Slimming
\cite{liu2017learning} baseline and achieve competitive performance among the concurrent
data-dependent and data-independent pruning approaches,
indicating the essential role of the inter-layer dependency in principled
filter pruning algorithms.

\section*{Acknowledgments}
This research was supported by Major Project for New Generation of AI under 
Grant No. 2018AAA0100400, NSFC (61922046), 
the national youth talent support program, 
and Tianjin Natural Science Foundation (18ZXZNGX00110).

\bibliographystyle{IEEEtran}
\bibliography{pruning}

% \newcommand{\AddPhoto}[1]{\includegraphics%
% [width=1in,height=1.25in,clip,keepaspectratio]{figures/authors/#1}}

% \begin{IEEEbiography}[\AddPhoto{kai}]{Kai Zhao}
% received his B.S. and M.S. from Shanghai University.
% He is currently a Ph.D. Candidate with the
% College of Computer Science, Nankai University, under the supervision of
% Prof. Ming-Ming Cheng.
% His research interests include statistical learning and computer vision.
% \end{IEEEbiography}

% \begin{IEEEbiography}[\AddPhoto{xyzhang}]{Xin-Yu Zhang}
% is an undergraduate student from the School of Mathematical 
% Sciences at Nankai University. 
% His research interests include computer vision and deep learning.
% \end{IEEEbiography}

% \begin{IEEEbiography}[\AddPhoto{hanqi}]{Qi Han}
% is a master student from the school of computer science at Nankai University, under the supervision of Prof. Ming-Ming Cheng. He received bachelor degree from Xidian University in 2019.
% His research interests include deep learning and computer vision.
% \end{IEEEbiography}

% \begin{IEEEbiography}[\AddPhoto{cmm}]{Ming-Ming Cheng}
% received his PhD degree
% from Tsinghua University in 2012. Then he did
% 2 years research fellow, with Prof. Philip Torr in
% Oxford. He is now a professor at Nankai University, leading the Media Computing Lab. His
% research interests includes computer graphics,
% computer vision, and image processing. He received research awards including ACM China
% Rising Star Award, IBM Global SUR Award,
% CCF-Intel Young Faculty Researcher Program,
% \etal .
% \end{IEEEbiography}

\vfill

\end{document}